\documentclass[conference]{IEEEtran}
\IEEEoverridecommandlockouts

\usepackage{cite}
\usepackage{amsmath,amssymb,amsfonts}
\usepackage{graphicx}
\usepackage{textcomp}
\usepackage{xcolor}
\usepackage{subcaption}
\usepackage{multirow}
\usepackage{hyperref}
\usepackage{balance}

\def\BibTeX{{\rm B\kern-.05em{\sc i\kern-.025em b}\kern-.08em
    T\kern-.1667em\lower.7ex\hbox{E}\kern-.125emX}}

\begin{document}

\title{GAIPAT -- Dataset on Human Gaze and Actions for Intent Prediction in Assembly Tasks
}

\author{
\IEEEauthorblockN{Maxence Grand}
\IEEEauthorblockA{\textit{Univ. Grenoble Alpes,} \\
\textit{CNRS, Grenoble INP, LIG,}\\
38000 Grenoble, France \\
Maxence.Grand@univ-grenoble-alpes.fr}
\and
\IEEEauthorblockN{Damien Pellier}
\IEEEauthorblockA{\textit{Univ. Grenoble Alpes,} \\
\textit{CNRS, Grenoble INP, LIG,}\\
38000 Grenoble, France \\
Damien.Pellier@univ-grenoble-alpes.fr}
\and
\IEEEauthorblockN{Francis Jambon}
\IEEEauthorblockA{\textit{Univ. Grenoble Alpes,} \\
\textit{CNRS, Grenoble INP, LIG,}\\
38000 Grenoble, France \\
Francis.Jambon@univ-grenoble-alpes.fr}
}

\maketitle

\begin{abstract}
The primary objective of the dataset is to provide a better understanding of the coupling between human actions and gaze in a shared working environment with a cobot, with the aim of significantly enhancing the efficiency and safety of human-cobot interactions. More broadly, by linking gaze patterns with physical actions, the dataset offers valuable insights into cognitive processes and attention dynamics in the context of assembly tasks.

The proposed dataset contains gaze and action data from approximately 80 participants, recorded during simulated industrial assembly tasks. The tasks were simulated using controlled scenarios in which participants manipulated educational building blocks. Gaze data was collected using two different eye-tracking setups --head-mounted and remote-- while participants worked in two positions: sitting and standing.
\end{abstract}

\begin{IEEEkeywords}
Human Robot Interaction, Human Intentions Prediction, Eye Tracking
\end{IEEEkeywords}

\section{Study Overview}

Initially, industrial robots operated in isolation from humans. However, in the past decade, the concept of "cobots" \cite{colgate1996cobots} --collaborative robots designed to work alongside humans-- has emerged. Cobots offer the advantage of reducing human-related constraints while improving performance, without replacing human workers \cite{peshkin1999cobots,fournier2024human}.

In these collaborative environments, cobots and humans work closely together, sharing tasks, spaces, and resources. Therefore, understanding how humans perceive the shared environment and the actions they perform is crucial for enhancing the efficiency and safety of human-cobot interactions. For example, if a cobot can anticipate that a human is about to pick up an object, it can adjust its movements to avoid interference. As human gaze often precedes physical actions \cite{johansson2001eye, mennie2007look}, gaze analysis becomes a promising tool for predicting future actions.

To explore these insights, we developed GAIPAT, a dataset to study the coupling between human gaze and actions in simulated industrial assembly tasks. Several existing datasets have also explored the link between gaze and actions, such as the Atari-HEAD dataset \cite{DBLP:conf/aaai/ZhangWLGMWZHB20}, RW4T \cite{DBLP:conf/hri/Orlov-SavkoQGNC24}, COLET \cite{DBLP:journals/cmpb/KtistakisSMTTFT22}, MoGaze \cite{DBLP:journals/ral/KratzerBMPTM21}, GIMO \cite{DBLP:conf/eccv/ZhengYMLYLLG22}, GTEA Gaze+ \cite{DBLP:journals/pami/LiLR23}, TIA \cite{DBLP:conf/cvpr/WeiLSZZ18}, EgoBody \cite{DBLP:conf/eccv/ZhangMZQKPBT22}). However, While Attari-HEAD, RW4T, COLET and GIMO track gaze in virtual and screen-based setting, only MoGaze, GTEA Gaze+, TIA and EgoBody track gaze on physical scene. A major limitation of these last datasets is that they primarily focus on daily life actions, making their experimental setups less relevant to industrial tasks. Moreover, these datasets typically use only a single device for gaze tracking. The GAIPAT dataset addresses this gap by including different configurations of eye-tracking devices --head-mounted and remote-- and operator positions --sitting and standing-- to explore various technical setups. Industrial assembly tasks are simulated with controlled scenarios in which operators manipulate educational building blocks.

\section{Methods}

We will now detail our approach to devising the dataset. Section \ref{sec:metod:setup} describes our experimental setup, including the environment where participants performed the assembly tasks, a comprehensive overview of all the sensors used, and the data recording procedure. Next, Section \ref{sec:method:participants} introduces the participants involved in the study. Section \ref{sec:method:procedure} outlines the experimental procedure. Finally, Section \ref{sec:method:data} covers the data collection and annotation processes.

\subsection{Experimental Setup}\label{sec:metod:setup}

\subsubsection*{Assembly Workspace}

\begin{figure}[t]
\centering
\includegraphics[width=.7\linewidth]{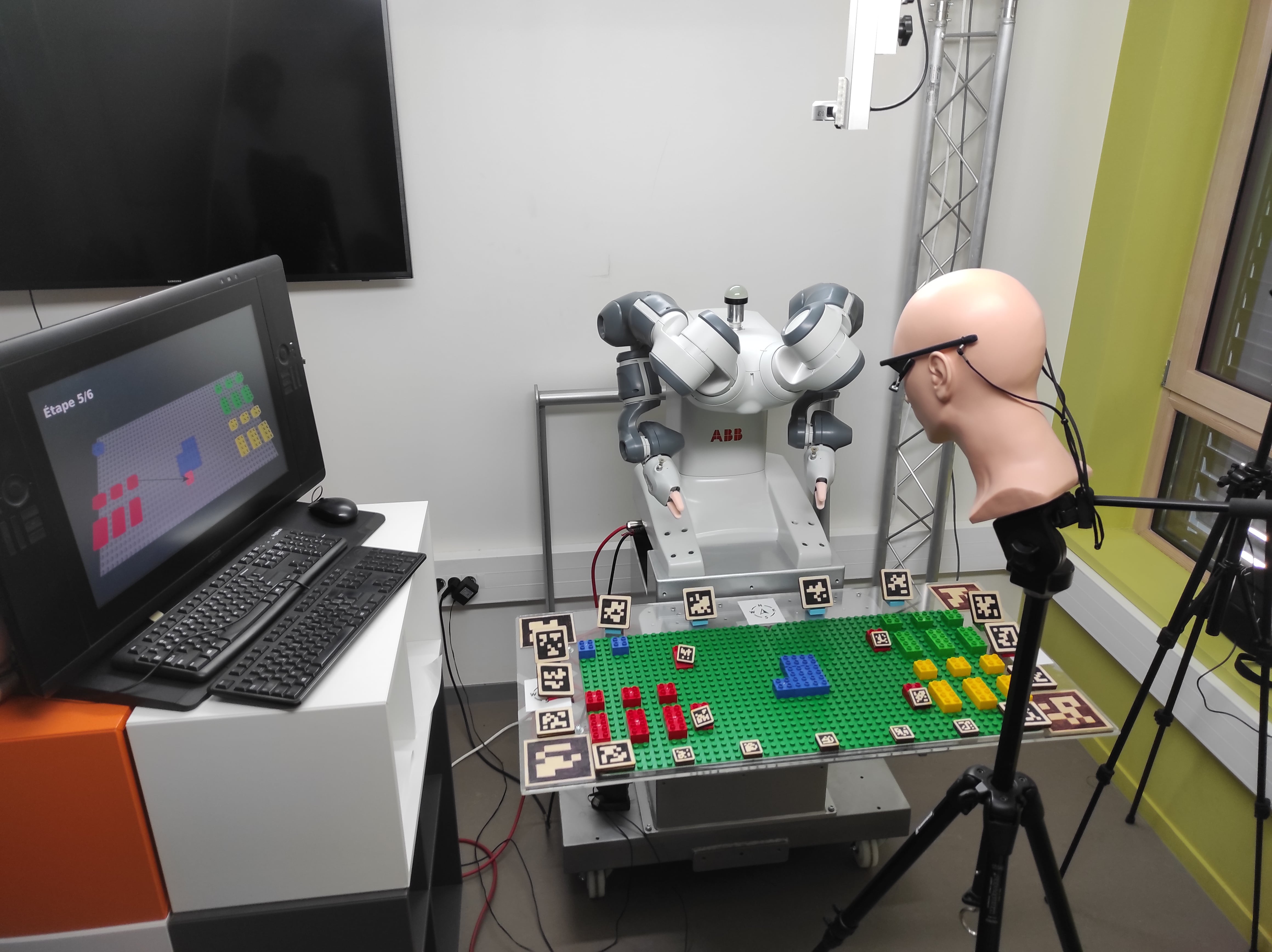}
\caption{Assembly Workspace}
\label{fig:method:assembly}
\end{figure}

We selected a workspace similar to those used in previous cobotics studies simulating industrial tasks \cite{hmedan2022adapting,fournier2024human}. The workspace (see Fig. \ref{fig:method:assembly}) consists of two main components: an assembly table and an instruction screen. The assembly table includes a green plate with a central assembly area and storage sections for construction blocks on either side. The blocks (Lego Duplo) are organized by color (blue, red, yellow, or green) and shape (cube or brick) to facilitate easy identification, mirroring industry-standard workspaces. Participants performed assembly tasks proposed by Younes et al. \cite{younes22} in both sitting and standing positions to cover a broad range of industrial scenarios. The table height was adjusted for each participant in accordance with the ISO 14738 standard \cite{ISO14738}. Participants were unfamiliar with the specific assemblies they needed to complete, which were communicated to them through an instruction screen. This screen, positioned next to the table, is operated via buttons located underneath to avoid disrupting the assembly process.

\subsubsection*{Eye-Tracking Configurations}

Two different eye-tracking configurations were used in the dataset to record human gaze:

\begin{enumerate}
\item {\em Remote configuration:} In this setup, devices are positioned at various locations within the participant’s workspace. We used the Fovio/FX3\footnote{\url{https://www.eyetracking.com/fx3-remote-eye-tracking/}} and the Tobii 4C\footnote{\url{https://help.tobii.com/hc/en-us/sections/360001811457-Tobii-Eye-Tracker-4C}}. The Fovio/FX3 tracked eye movements on the table. We selected this remote eye-tracker for its ability to monitor horizontal areas, unlike most devices designed for vertical screens. As recommended by the documentation, it was positioned at the bottom of the table to avoid eyebrow occlusion and ensure effective calibration. However, user arm movements may obstruct tracking. Although multiple Fovio/FX3 units could be used side by side, our setup accommodated only one. Data from the Fovio/FX3 were recorded using Eyeworks software\footnote{\url{https://www.eyetracking.com/eyeworks-software/}}. The Tobii 4C --upgraded with an 'analytical use' license-- tracked eye movements on the instruction screen. Initially designed for gaming, the Tobii 4C is less accurate than other Tobii models but is more tolerant of head movement and supports a greater distance between the tracker and participants. Tobii 4C data were collected using custom Python scripts.

\item {\em Head-mounted configuration:} In this setup, participants wore Pupil Labs Core eye-tracking glasses\footnote{\url{https://pupil-labs.com/products/core}} \cite{pupil-lab}, equipped with a scene camera above the right eye and two infrared cameras to measure eyes movements. We used AprilTag markers from the 'tag36h11' family \cite{apriltag} to map the assembly workspace to the scene camera. Data were recorded using the Pupil Labs Capture software\footnote{\url{https://docs.pupil-labs.com/core/software/pupil-capture/}}.

\end{enumerate}

\subsection{Participants}\label{sec:method:participants}

We recruited 80 psychology students who volunteered to participate in the study, divided into four groups of 20 based on the remote and head-mounted configurations, with participants either standing or sitting. No personal data (e.g., gender, age, health) was collected. Participants were compensated with experience points for their semester exams. The inclusion criteria required participants to be able to read and understand instructions in French and physically move construction blocks. Since the tasks could be performed while either sitting or standing, the ability to stand was not a criterion for inclusion.

\subsection{Procedure}\label{sec:method:procedure}

\begin{figure}[t]
\centering
\begin{subfigure}[t]{.3\hsize}
\centering
\includegraphics[width=.8\linewidth]{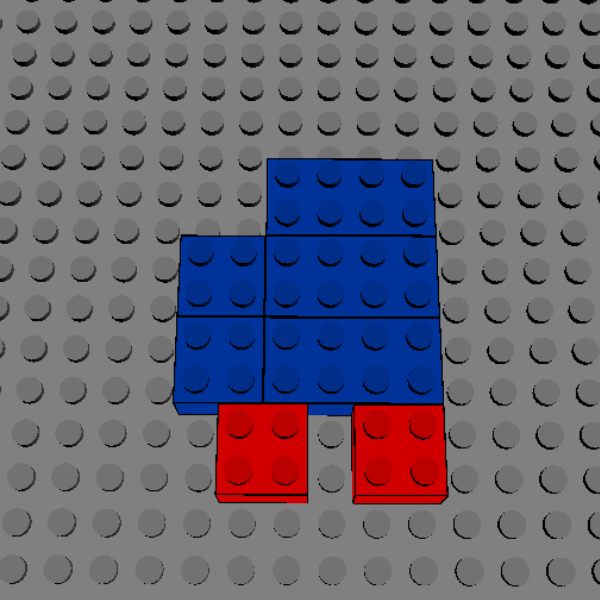}
\caption{Car}
\label{fig:method:task:figure:car}
\end{subfigure}
\begin{subfigure}[t]{.3\hsize}
\centering
\includegraphics[width=.8\linewidth]{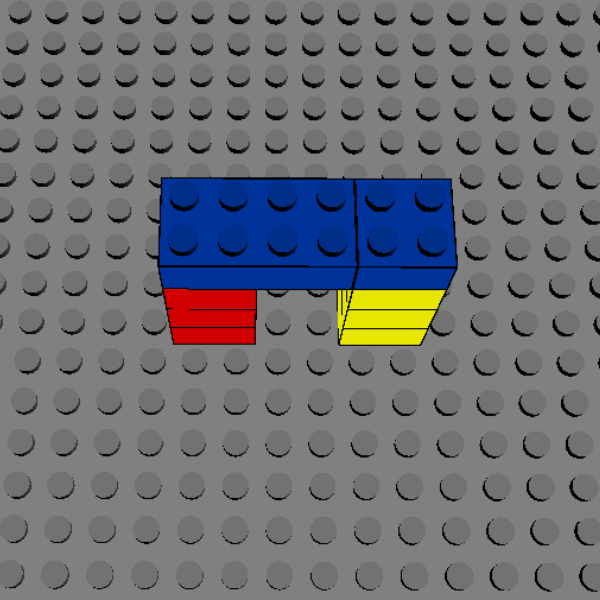}
\caption{Towers and Bridge}
\label{fig:method:task:figure:tb}
\end{subfigure}
\begin{subfigure}[t]{.3\hsize}
\centering
\includegraphics[width=.8\linewidth]{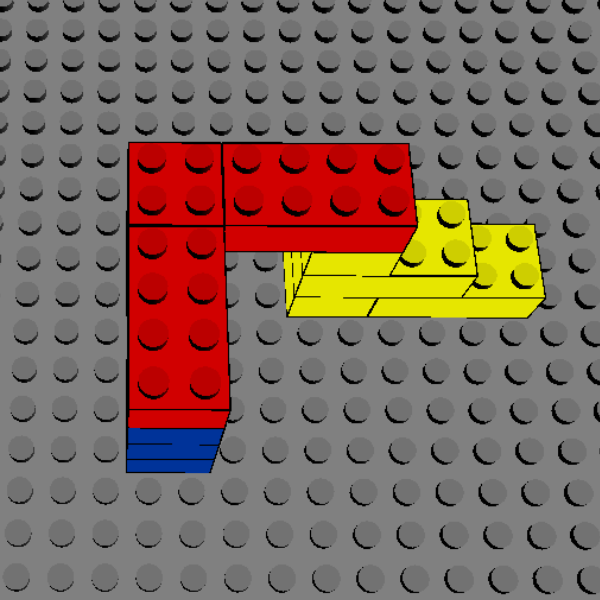}
\caption{Tower, Stairs and Bridge}
\label{fig:method:task:figure:tsb}
\end{subfigure}
\caption{Assembly Tasks -- Examples of figures}
\label{fig:method:tasks}
\end{figure}

Fig. \ref{fig:method:tasks} shows an excerpt from the sequence of figures that participants must assemble.

\begin{figure}[t]
\centering
\begin{subfigure}[t]{.45\hsize}
\centering
\includegraphics[width=\linewidth]{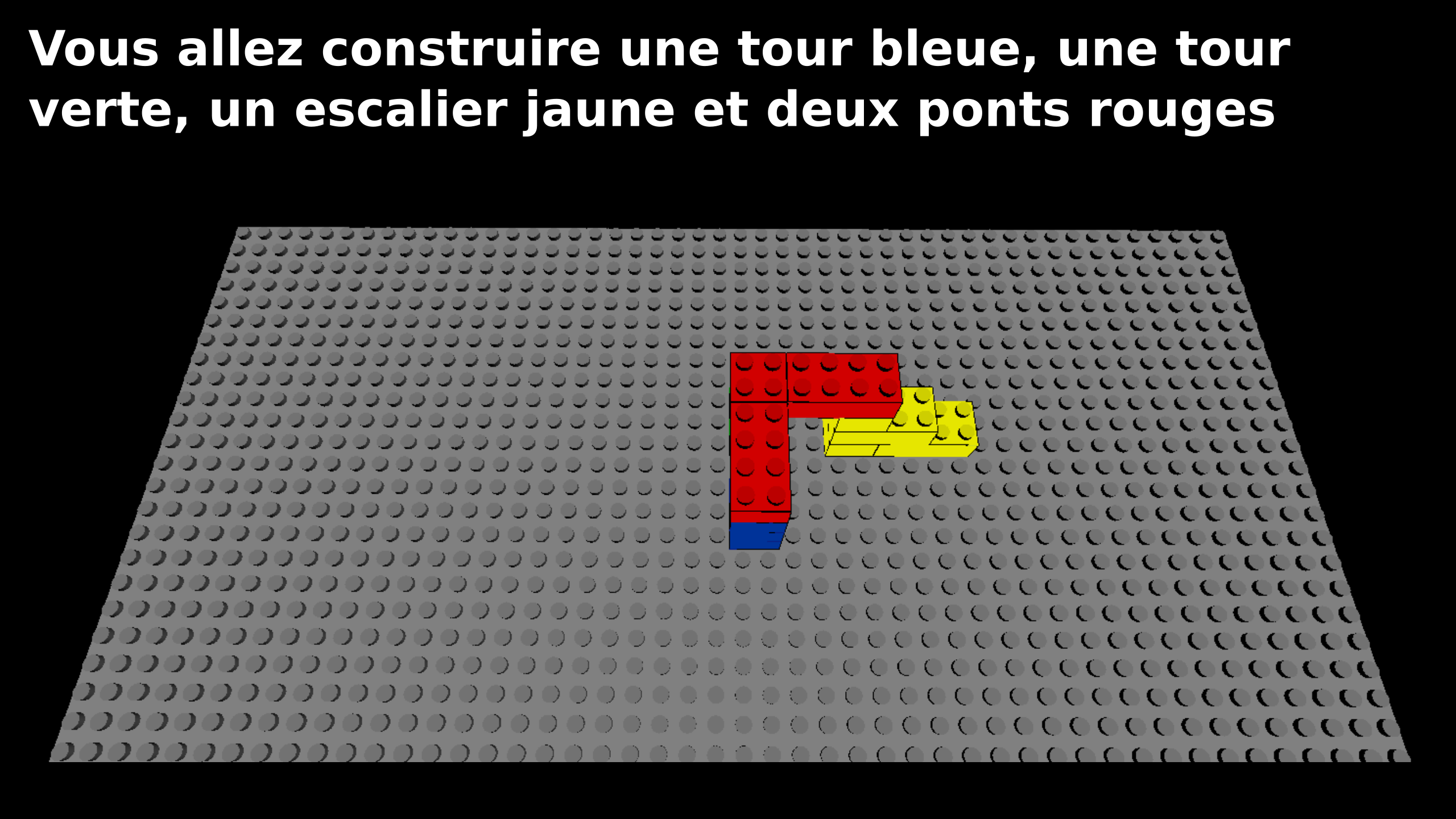}
\caption{The initial instruction.\\{\it Translation: "You'll build a blue tower, a green tower, yellow stairs, and two red bridges."}}
\label{fig:method:instructions:initial}
\end{subfigure}
\begin{subfigure}[t]{.45\hsize}
\centering
\includegraphics[width=\linewidth]{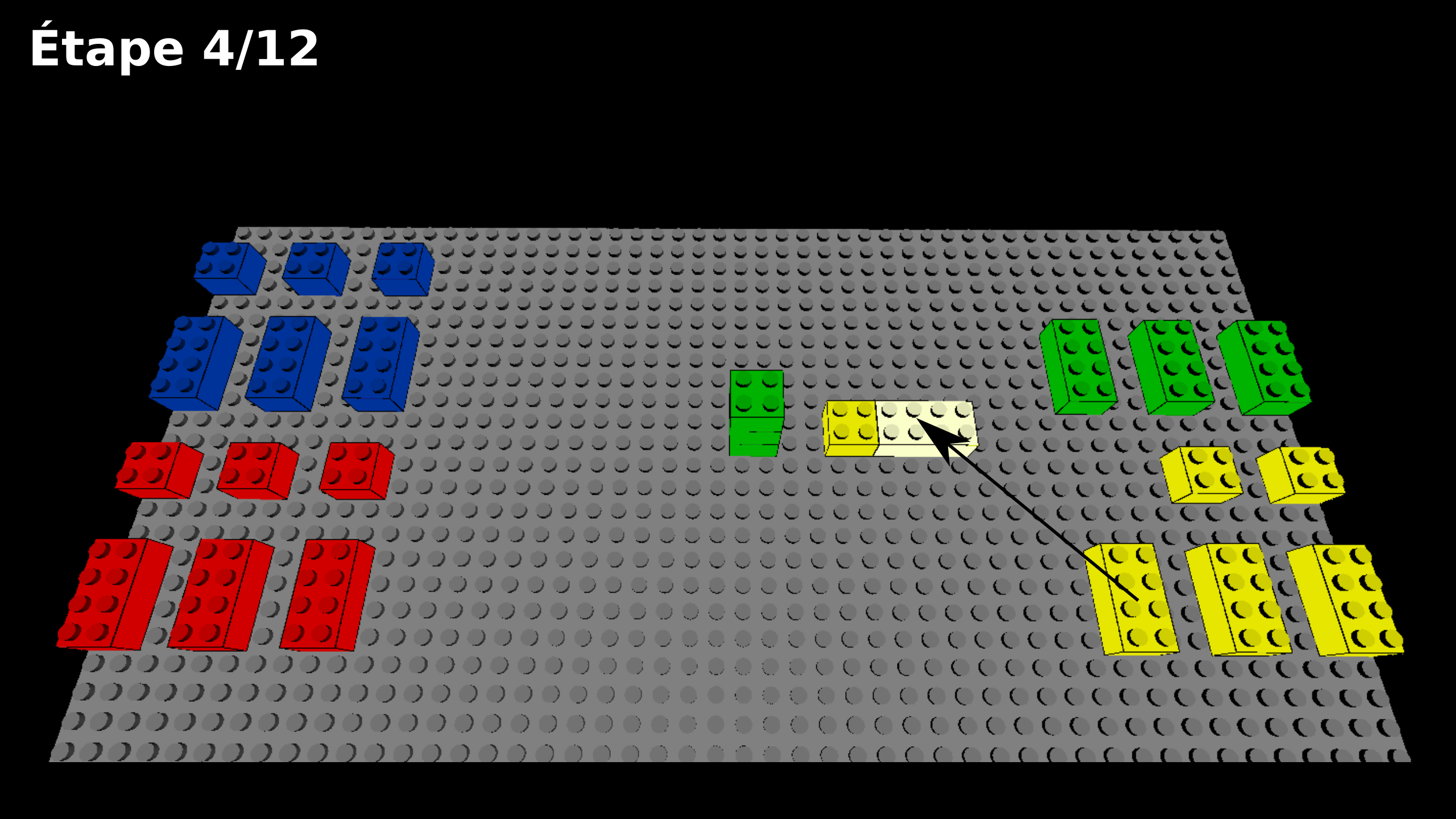}
\caption{The step-by-step instruction.\\{\it Translation: "Step 4/12."}}
\label{fig:method:instructions:step}
\end{subfigure}
\caption{Assembly Tasks -- Instructions}
\label{fig:method:instructions}
\end{figure}

Participants performed assembly tasks while their eyes and hands movements were tracked by various sensors. Fig. \ref{fig:method:tasks} shows an excerpt from the sequence of figures that participants must assemble. The difficulty of the figures was varied by adjusting the number of construction blocks used, as well as by incorporating both 2D and 3D figures. Instructions for assembling these figures were displayed on a separate screen (see Fig. \ref{fig:method:instructions}). The experimental procedure was the same for all participants and figures:
\begin{enumerate}
\item The eye-tracking device was calibrated and the calibration validated. If calibration quality was insufficient, this step was repeated. Manufacturer's calibration and validation guidelines were followed. In cases where calibration could not be achieved, the experiment continued, but data were not recorded. For each task, eye movements were measured both on the assembly table and the screen. 
\item The participant's first instruction appeared on the screen, providing the context and showing the figure to be assembled (see Fig. \ref{fig:method:instructions:initial}). The experimenter placed the first construction block, and the remaining blocks were placed by the participant. 
\item At each step, a visual instruction indicated which construction block to pick up and where to place it in the assembly (see Fig. \ref{fig:method:instructions:step}). The instructions were structured such that, for each step, the block to be moved and its destination were highlighted. Additionally, to eliminate any ambiguity, an arrow indicated where to move the block. Each figure required six to twelve steps to complete. Participants advanced through each step by pressing a button located under the table.
\end{enumerate}

The entire procedure take approximately 5 minutes per figure, and 30 minutes for all figures. Taking into account the welcoming of the participant, the collection of consent and the explanation of the experiment, the experiment last around 50 minutes per participant.

\begin{figure}[t]
    \centering
    \includegraphics[width=.8\linewidth]{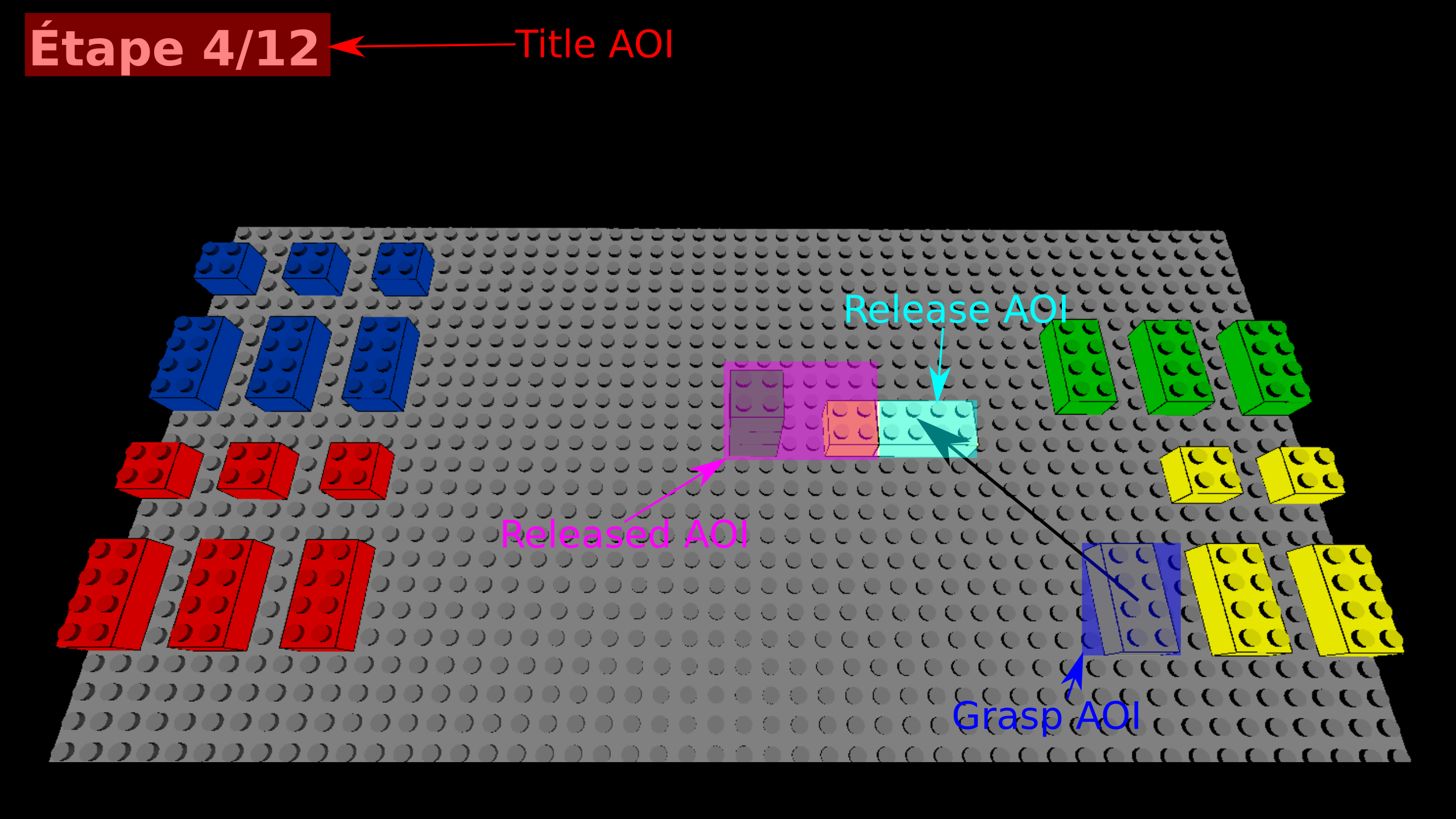}
    \caption{Example of Screen AOIs}
    \label{fig:method:annotation:screen_aoi}
\end{figure}

\subsection{Data Collection and Annotation}\label{sec:method:data}

We collected raw data from each eye-tracker, along with video recordings of the participants’ hand movements and system timestamps. To ensure privacy, only the participants' hands were visible in the videos, and any distinctive features like tattoos or jewelry were concealed or removed. Furthermore, the videos themselves were excluded from the dataset, with only the annotations retained. The raw data and video annotations were synchronized using video timestamps, linking the eye-tracking data to the participants' hand gestures. For each participant, a 30-minute video was recorded.

There are two types of annotations. First, event annotations were manually made from the videos to mark when a specific action was performed by the participant. Second, each instruction slide was annotated with Areas of Interest (AOIs), representing the different zones on the slide.

\subsubsection*{Events Annotation}

There are two types of events to annotate: those occurring on the assembly table and those on the instruction screen.

For the table events, participants performed several actions, primarily "picking up" and "placing" construction blocks. Each action involved multiple sub-actions and gestures, such as reading instructions, locating, grasping, releasing, and verifying instructions. Defining the start and end times of these actions can be imprecise, and extraneous gestures and hesitations further complicated the annotation process. Therefore, we chose to annotate only unambiguous atomic events, defined as follows: a \textit{"grasp"} event is when the participant’s fingers make contact with the construction block; a \textit{"release"} event is when the participant’s fingers release the block without intending to subsequently move it. Additionally, \textit{"start"} (when figure assembly begins) and \textit{"end"} (when figure assembly ends) events were annotated to mark the beginning and conclusion of the assembly task.

For the screen events, buttons under the table allowed participants to navigate through the instructions. \textit{"Next"} (and \textit{"previous"}) events were annotated to indicate when the participant pressed the button to move forward (or backward) in the instructions. As with table events, \textit{"start"} and \textit{"end"} events were also annotated to denote the beginning and end of the instruction sequence.

\subsubsection*{Screen Area of Interests Annotation}

In addition to annotating events, we also annotated the instruction slides. Unlike the assembly table, where the focus is primarily on the positions of the construction blocks, instruction slides contain additional information, such as the "title instruction," which provides context and progress for the assembly, and the designated areas where construction blocks should be moved. To effectively capture this information, we decompose and annotate the instructions using Areas of Interest (AOIs). The annotations are organized as follows (see Fig. \ref{fig:method:annotation:screen_aoi}):
\begin{itemize}
\item \textit{"Title AOI"}: this area, located at the top of the instruction, includes contextual information and progress indicators for the assembly task.
\item \textit{"Grasp AOI"}: this area highlights where the construction block to be moved is currently located.
\item \textit{"Release AOI"}: this area specifies where the construction block should be placed.
\item \textit{"Released AOI"}: this area indicates where previous construction blocks have already been placed.
\end{itemize}
By using these AOIs, we aim to provide a clear and detailed understanding of the instructions and their relevance to the assembly task.

\section{Dataset}

The dataset is organized into two main categories: {\em Setup Data}, which includes information on participant characteristics, block descriptions, and instruction details collected before the assembly tasks, and {\em Participant Data}, which captures detailed records of the participants' actions and gaze measurements during the figure assembly process.

\subsection{Setup Data}

Setup data, located in the \texttt{setup} directory, consists of five files:

\begin{enumerate}

\item {\em blocks.csv:} This file describes the blocks used in the assembly tasks, including each block's ID, color (blue, red, green, yellow), and shape (cube or brick).

\item {\em instructions\_\{figure\_name\}.csv:} This file provides step-by-step instructions for assembling each figure, specifying the block ID to be moved, its original position, and its destination position, described by the coordinates of its corners, and the level where it is supposed to be placed.

\item {\em slides\_\{figure\_name\}.csv:} This file details the slides used during the assembly process, including the slide number and the positions of its Areas of Interest (AOIs) such as the title, block to grasp, block destination, figure area, and arrow. Each AOI is defined by the coordinates of its corners.

\item {\em participants.csv:} This file contains participants information, including IDs, setups (remote or head-mounted eye-trackers), positions (sitting or standing), calibration scores (no issue, slight issues, severe issues, or impossible), and whether data was recorded for each figure. Calibration scores are defined as follows: 
\begin{itemize} 
\item {\em No issue:} No problems were detected during the calibration of the eye-trackers. 
\item {\em Slight issues:} Some targets during calibration were not successfully completed on the first attempt. Calibration and/or validation had to be repeated but succeeded without the participant moving his/her head or the calibration targets being repositioned. 
\item {\em Severe issues:} It was necessary to reposition the calibration targets and/or allow the participant to move her/his head. 
\item {\em Impossible:} Calibration of the eye-trackers was not possible. Consequently, no data was recorded. 
\end{itemize}

\item {\em glasses.csv:} This file provides information on the distribution of glasses and contact lenses among participants. It categorizes the data by setup type and position, and includes counts of participants who wore glasses versus those who did not. To maintain participant anonymity, details about the wearing of glasses are not linked to individual participant IDs, as such information could be identifying.

\end{enumerate}

\subsection{Participant Data}

The participant data, located in the \verb|participant/[participant_id]| directory, details all the events that occurred during the assembly of the figures. This data provides a comprehensive account of the assembly process and includes all eye tracker measurements, such as gaze points, pupil sizes, and the confidence levels in the accuracy of these measurements. The data is separated according to the specific figures being assembled in 3 files:
\begin{enumerate}
\item {\em events.csv:} This file records events that occurred on the table and screen during the assembly process. For the table, it includes events such as start, grasp, release, and end, along with the associated block IDs. For the screen, it includes events such as start, next, previous, and end.
\item {\em states.csv:} This file documents the state of the table and screen at various timestamps. For the table, it includes the positions and status of each block (whether it is held by the participant). For the screen, it records the slide number being displayed.
\item {\em gazepoints.csv:} This file captures the gaze points on both the table and the screen during the assembly tasks. The data is standardized across different eye-trackers, retaining only the "both eyes" gaze points. It includes timestamps and (x, y) coordinates of the gaze points. Also, as many methods exist to detect fixations (e.g., dispersion-based \cite{dispersion}, ML-based \cite{ml-based}, velocity-based \cite{nystrom2010adaptive}), we chose to retain raw gaze point data and did not include fixation calculations in the dataset.
\end{enumerate}

\section{Usage Notes}

\subsubsection*{Use Cases} The dataset can be used in the following use cases: (1) {\em Human Intent Prediction:} it helps to explain and anticipate human actions by analyzing gaze patterns, offering thus an insight into the cognitive processes behind physical gestures in the context of assembly tasks; (2) {\em Generative Gaze Model:} it provides a baseline for creating cobot gaze models immitating human gaze behavior, improviving human anticipation of cobots' actions; (3) {\em Enhanced Human-Cobot Collaboration:} by determining the distribution of human attention during assembly tasks, the dataset contribues to improving the efficiency and safety of human-cobot collaboration.

\subsubsection*{Versatility} The dataset is versatile and covers a range of experimental conditions: (1) {\em Diverse Tracking Technologies:} it includes data from both remote and head-mounted eye-trackers, allowing for comparisons between different eye-tracking technologies; (2) {\em Posture Variations:} participants completed tasks in both sitting and standing positions, offering insights into how posture influences gaze behavior and (3) {\em Real-World Relevance:} the experimental setup closely mirrors those used in existing human-robot interaction studies, ensuring relevance to practical collaborative environments.

\subsubsection*{Limitations} The dataset has some limitations: (1) {\em Simplistic Tasks:} the figures assembled were relatively simple, and the assembly process was fully guided by instructions, which may not fully represent the complexity of real-world tasks and (2) {\em Tracking Technology:} only infrared-based eye-trackers were used, which may limit the generalizability of the results to other eye-tracking technologies, as webcam based eye-tracking.

\subsubsection*{Repositories and Documentation} The repository, which includes comprehensive documentation and the ACM datasheet \cite{DBLP:journals/cacm/GebruMVVWDC21}, is publicly available\footnote{\url{https://gricad-gitlab.univ-grenoble-alpes.fr/eyesofcobot/gaipat}}. This documentation has been carefully designed to assist external users in understanding and utilizing the dataset effectively. It provides detailed explanations of the dataset structure, data collection methods, and usage guidelines. In addition, the raw data collected during the experiment, which was used to generate the dataset, is also publicly. This raw data is accompanied by thorough documentation to support its use and ensure clarity in data interpretation.

\subsubsection*{Ethical Statement} The experimental protocol for creating the dataset was reviewed and approved by  CERGA, the Grenoble Alpes University ethics committee, and received the favorable opinion CERGA-Avis-2023-32. 

\section{Acknowledgements}

This work is supported by the French National Research Agency in the framework of the "Investissements d’avenir” program (ANR-15-IDEX-02).

\bibliographystyle{IEEEtran}
\balance
\bibliography{IEEEabrv,references.bib}

\end{document}